\documentclass{article}
\usepackage{ijcai27}
\usepackage{xcolor}

\usepackage{times}
\usepackage{soul}
\usepackage{url}
\usepackage[hidelinks]{hyperref}
\usepackage[utf8]{inputenc}
\usepackage[small]{caption}
\usepackage{graphicx}
\usepackage{amsmath}
\usepackage{amsthm}
\usepackage{booktabs}
\usepackage{algorithm}
\usepackage{algorithmic}
\usepackage[switch]{lineno}
\usepackage{array}
\usepackage{tabularx}
\usepackage{amssymb}
\usepackage{multirow}
\newcounter{validationstep}

\newcommand{\stepheading}[1]{%
    \par\addvspace{0.6\baselineskip}%
    \refstepcounter{validationstep}%
    \noindent\textbf{(\arabic{validationstep})\ #1}%
    \par\smallskip
}

\title{SIRCA: Root Cause Attribution in LLM Agent Trajectories via Static Interventions}

\author{
Kun Jiang \and
Han Lin \and
Xingyu Liao \and
Guang Yang \and
Mengjie Zhang \and
Sijie Wu \and
Kai Zhang \and
Xingyu Yan \and
Bowen Liu \and
Jiantao Ye \and
Qitao Song \and
Xubin Li \and
Xin Chen\\
\affiliations
Huawei Technologies Co., Ltd., AI Data Department, ICT GTS, Shanghai, China\\
\emails
lixubin@huawei.com
}

\begin{document}

\maketitle
\begingroup
\renewcommand{\thefootnote}{}
\footnotetext{The source code and dataset are publicly available at \url{https://github.com/lh1293151824-max/SIRCA}}
\endgroup

\begin{abstract}
Identifying the root causes of failures in large language model (LLM) agents is difficult because errors may propagate across multiple trajectory steps, thereby obscuring their true origins. User feedback can often point to specific problems in agent behavior, providing direct evidence for tracing error origins. To this end, we propose Static Intervention for Root Cause Attribution (SIRCA), a method that leverages user feedback to localize root causes in agent trajectories without replaying agent execution. SIRCA uses negative user feedback as an attribution anchor, constructing an attribution graph to model error origins and their propagation processes. For trajectories without negative user feedback, it instead constructs alternative anchors from task failure information. SIRCA then determines the final root causes through LLM-based evaluation of static intervention repair proposals. On the Algorithm Generated (AG) and Hand Crafted (HC) datasets of Who\&When, SIRCA achieves relative improvements of $36.59\%$ and $13.51\%$, respectively, in root cause localization accuracy over the strongest baseline on each dataset. On the multi root cause wireless agent dataset with user feedback, SIRCA improves trajectory averaged $F_1$ and recall by $9.58\%$ and $32.52\%$, respectively, relative to the strongest baseline. In addition, we release the wireless agent dataset and the SIRCA source code to support future research.
\end{abstract}


\section{Introduction}

Large language model (LLM) agents execute tasks through consecutive steps such as reasoning, tool interactions, and responses \cite{yao2023react,zeng2024agenttuning,qin2024toolllm}. Their execution trajectories provide important evidence for diagnosing failures and identifying critical steps that lead to them \cite{zhu2025llm,zhang2025agent}. However, an error may originate at one step and propagate through subsequent actions, becoming apparent only in a later response or task outcome \cite{zhang2026longrca,li2026towards}. For example, an incorrect parameter assumption may be repeatedly reused across multiple calculations before being exposed in the final recommendation. Therefore, accurately localizing the root causes of agent failures requires a thorough analysis of the relationships between trajectory steps and subsequent failure manifestations.

Recent studies on root cause localization in agent trajectories have primarily focused on the relationship between trajectory steps and task failures, identifying critical steps \cite{ge2026spectrum,banerjee2025did,xia2026broke}, modeling potential causal relationships \cite{wang2026dcfa,wang2026flat}, and analyzing error propagation \cite{ma2025demystifying,rafi2026falat} to localize the key factors that lead to task failures. These studies show that systematically analyzing structured trajectories and the causal propagation relationships within them can provide a deeper understanding of how agent failures arise. Building on these research directions, this work further focuses on how to determine the final root causes from a trajectory based on evidence related to specific issues.

During real world interactions with LLM agents, users may correct an erroneous value, reject an unsatisfactory response, or point out an omitted action \cite{yan2024refutebench,tandon2022learning}. Such feedback can reveal concrete problems encountered during task execution and provide issue specific evidence for failure diagnosis \cite{zhu2026autolibra}. Prior studies have found that user feedback can provide effective guidance for improving agent behavior \cite{cai2026ievoagent,petrak2024learning,hancock2019learning}. Furthermore, recent work \cite{zhao2026trace} has used user dissatisfaction signals in trajectories to diagnose errors in agent context sources, demonstrating the important value of user feedback in interaction trajectories for analyzing failures in agent trajectories. Although user feedback provides issue specific clues, accurately tracing the reported problems to their root causes remains challenging. LLM agents generate outputs autoregressively according to conditional probability distributions over tokens given the preceding context \cite{radford2019language}, allowing information from earlier trajectory steps to influence subsequent outputs \cite{li2025generator}. A problem identified by the user may therefore reflect the combined influence of multiple steps \cite{shah2026causal} that introduce, inherit, or manifest an error \cite{rafi2026falat}, making their individual contributions difficult to disentangle and the error origin difficult to identify.

To address these challenges, this work investigates how to leverage evidence provided by negative user feedback to model error origins and their propagation processes, thereby localizing the root causes of failures in agent trajectories. Our main contributions are summarized as follows:

\begin{itemize}
\item We propose Static Intervention for
Root Cause Attribution (SIRCA), a method that leverages negative user feedback to localize root causes in agent trajectories without replaying agent execution. Based on structured trajectory steps, SIRCA constructs an attribution graph anchored in negative user feedback to model error origins and their propagation processes, and performs root cause localization through LLM-based evaluation of static intervention repair proposals.

\item We release a wireless agent dataset collected from telecommunications network operations and wireless network optimization scenarios, with explicit negative user feedback and step level root cause annotations, together with the SIRCA source code.
\end{itemize}

\section{Related Work}
\subsection{Agent Failure Attribution Analysis} 
Existing studies have explored various approaches to root cause analysis in agent trajectories, typically using task failure outcomes and trajectory analysis methods such as step-wise inspection \cite{zhang2025agent,zhu2025llm}, consistency verification \cite{xia2026broke}, and hierarchical attribution \cite{banerjee2025did} to identify critical error steps. These studies can localize root causes in agent trajectories to some extent, but their analysis of the relationship between trajectory steps and task failures remains limited. To address this limitation, some studies further apply causal interventions to trajectory steps \cite{ma2026dover,bonagiri2026causalflow}, examining their effects on the final task outcome to localize or validate the root causes of failure. However, focusing only on the effects of trajectory steps on the final outcome often makes it difficult to distinguish error origin steps from subsequent propagation steps. To address this issue, recent studies have begun modeling error propagation relationships among trajectory steps \cite{wang2026flat,li2026towards,rafi2026falat}, thereby enabling more accurate localization of the corresponding root causes.

\subsection{User Feedback for Failure Attribution} 

In general, task failure information only reflects the overall execution outcome and provides limited evidence about specific problems that arise within the trajectory \cite{madvil2026holistic,zhuge2025agent}. In contrast, negative user feedback can more directly identify problems in agent behavior and provide evidence associated with specific error manifestations, offering clearer clues for tracing their underlying origins \cite{yang2022toward,mahato2024exploring,liu2025user}. Prior studies have shown that user feedback can provide effective guidance for improving agent behavior. For example, some studies use feedback generated during interactions as training signals to improve model response quality \cite{don2024naturally,hancock2019learning}, while others leverage user feedback to enhance the model's ability to adapt to user preferences \cite{tucker2024coactive,cai2026ievoagent}. Building on these studies, Zhao et al.~\shortcite{zhao2026trace} further leverage user dissatisfaction signals in trajectories to identify potentially erroneous trajectory context sources and accordingly generate corresponding repair suggestions.

Overall, existing studies on user feedback for LLM agents have primarily focused on improving agent behavior and repairing potential errors in trajectory context, while systematic research on root cause localization in agent trajectories using user feedback remains limited.

\section{Method}

\begin{figure*}[h]
    \centering
    \includegraphics[width=\textwidth]{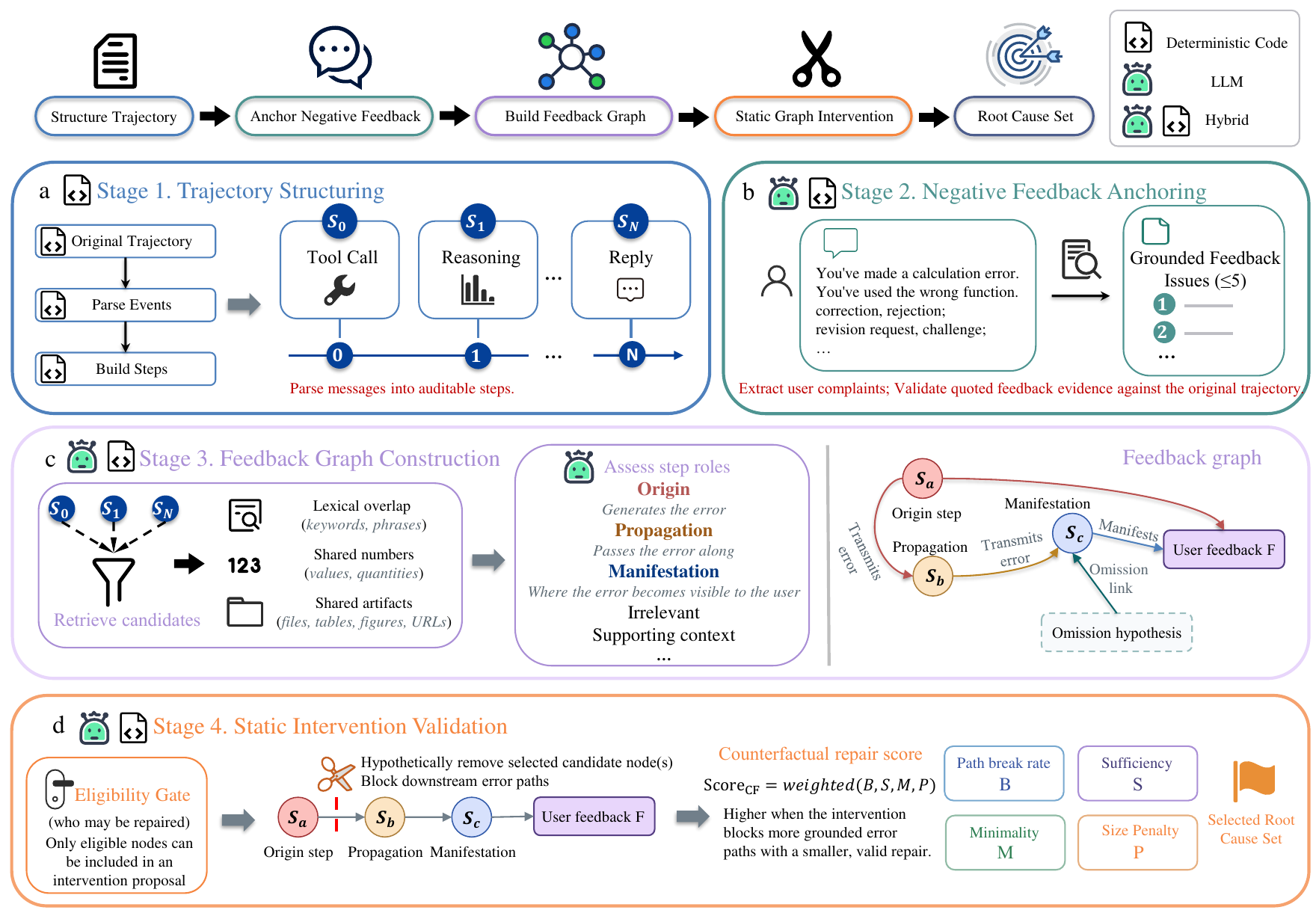}
    \caption{Overview of Static Intervention for Root Cause Attribution.}
    \label{fig:overview_sirca}
\end{figure*}

As shown in Fig.~\ref{fig:overview_sirca}, SIRCA combines LLM-based reasoning with deterministic rules to localize root causes of failures in agent trajectories. The overall framework consists of four interconnected stages: Trajectory Structuring, Negative Feedback Anchoring, Feedback Graph Construction, and Static Intervention Validation. In addition, all LLM prompts used in this work are publicly available in our open source repository.

\subsection{Trajectory Structuring}

Given a chronologically recorded trajectory of agent and user interactions, SIRCA parses agent reasoning, replies, user messages, and tool interactions into standardized events, then converts them into structured steps using deterministic rules. Specifically, nonempty agent reasoning and replies form reasoning and reply steps, respectively; user messages after the first round of interaction form candidate feedback steps; and each tool call is merged with the first subsequent unmatched tool observation to form a tool interaction step.

SIRCA indexes the steps in their original order of appearance and records their action summaries, available observation summaries, and neighboring context. This yields a step sequence $\mathcal{S}$ for subsequent evidence retrieval and semantic assessment:
\begin{equation}
\mathcal{S}=(S_0,S_1,\ldots,S_{N-1}).
\label{eq:structured-step-sequence}
\end{equation}
Here, $N$ is the total number of trajectory steps, and $j\in\{0,\ldots,N-1\}$ is the sequential index of step $S_j$.

\subsection{Negative Feedback Anchoring}

SIRCA selects user feedback steps from the structured step sequence $\mathcal S$ to form the set
\begin{equation}
\mathcal F^{\mathrm{cand}}
=
\left\{
S_j\in\mathcal S
\;\middle|\;
\operatorname{type}(S_j)=\texttt{user\_feedback}
\right\}
\end{equation}


Here, $\texttt{user\_feedback}$ is the structural label for user feedback steps. SIRCA then uses the LLM-based feedback classification model $\mathcal J$ to semantically assess the candidate steps using the trajectory context $\mathcal{S}^{(2)}$, which includes the trajectory up to and including the first agent step following the last candidate feedback step. Based on this context, $\mathcal J$ identifies user messages that point out agent errors and generates candidate negative feedback records $\hat F_i$. For trajectories without explicit user feedback, SIRCA appends a synthetic message derived from the available task failure information to the end of the trajectory as a substitute for negative user feedback, and further constructs an attribution anchor with the same structure as a user feedback record for subsequent root cause analysis.

SIRCA validates the candidate records $\hat F_i$ against the original trajectory to check their grounding in trajectory evidence. The validation procedure is detailed in Appendix~\ref{sec:Negative-Feedback Validation}. The validated records $F_i$ are then sorted by feedback step index in ascending order, and at most five are retained to form the negative feedback list $\mathcal F$:
\begin{equation}
\mathcal F=(F_1,\ldots,F_K),\quad K\leq5.
\end{equation}

The $i$th validated record is represented as
\begin{equation}
F_i=(f_i^*,y_i,t_i,q_i,M_i^*,E_i^*),
\quad q_i=(d_i,o_i,x_i).
\end{equation}

Here, $f_i^*$ is the validated negative feedback step index; $y_i$ and $t_i$ denote the feedback intent and issue type, respectively; $d_i$, $o_i$, and $x_i$ denote the issue description, erroneous content, and the expected response, respectively; and $M_i^*$ and $E_i^*$ denote the validated set of error manifestation step indices and the validated list of textual references provided by $\mathcal J$, respectively. 

\subsection{Feedback Graph Construction}

\subsubsection{Root Cause Candidate Step Retrieval}
\label{sec:feedback-driven-candidate-retrieval}

For each negative feedback record $F_i$, SIRCA restricts the corresponding root cause search space $\mathcal{A}_i^{-}$ to agent steps preceding the feedback step. To measure the relevance of each step in the root cause search space
$\mathcal A_i^{-}$ to negative feedback $F_i$, SIRCA combines the feedback
issue information $q_i=(d_i,o_i,x_i)$ with quotations from the validated
textual references $E_i^*$ to form the query text $u_i$. It also combines
the action, observation, and preceding context information of step $S_j$
into the step representation $\xi_j$. SIRCA then computes the retrieval
score using lexical and numerical overlap between $u_i$ and $\xi_j$,
together with file and tool identifiers shared by $S_j$ and the error
manifestation steps in $M_i^*$:
\begin{equation}
\begin{aligned}
\rho_{ij}
={}& L(u_i,\xi_j)
+\min\!\left(0.75,0.2n_{ij}^{\mathrm{num}}\right)\\
&+\min\!\left(0.75,0.25n_{ij}^{\mathrm{art}}\right).
\end{aligned}
\end{equation}
Here, $L(\cdot,\cdot)$ denotes the lexical overlap ratio, $n_{ij}^{\mathrm{num}}$
denotes the number of shared numerical values, and
$n_{ij}^{\mathrm{art}}$ denotes the number of file and tool identifiers
shared by $S_j$ and the error manifestation steps.

After obtaining $\rho_{ij}$, SIRCA first adds the error manifestation steps in $M_i^*$ that belong to $\mathcal A_i^{-}$ to the root cause candidate set. It then adds other steps in descending order of $\rho_{ij}$, retaining at most eight steps. To supplement potential root causes that may have been missed, SIRCA constructs a stratified candidate pool from the early, middle, and late portions of the trajectory. The semantic retrieval model $\mathcal J_{\mathrm{SR}}$ then selects steps from this pool that have not yet been included in the root cause candidate set and are relevant to the current negative feedback, and adds them to the set. Together, the two retrieval methods yield the final root cause candidate set
\begin{equation}
\mathcal R_i^{-}\subseteq\mathcal A_i^{-}.
\end{equation}

\subsubsection{Root Cause Candidate Step Role Assessment}

SIRCA applies the step assessment model $\mathcal J_{\mathrm{step}}$ to evaluate the role of each step in $\mathcal R_i^{-}$ with respect to negative feedback $F_i$ and assigns it a role label $\ell_{ij}$, thereby better distinguishing error origin, propagation, manifestation, and other cases. The roles are defined in Table~\ref{tab:stage3-role-definitions}, and detailed assessment examples are provided in Appendix~\ref{app:candidate-step-role-assessment}.

\begin{table}[h]
    \centering
    \small
    \setlength{\tabcolsep}{3pt}
    \renewcommand{\arraystretch}{1.1}
    \begin{tabularx}{\columnwidth}{
        @{}
        >{\raggedright\arraybackslash}m{0.29\columnwidth}
        >{\raggedright\arraybackslash}X
        @{}
    }
        \toprule
        \textbf{Role $\ell_{ij}$} & \textbf{Definition} \\
        \midrule
        \textbf{Error Origin}
        & First introduces the error that ultimately leads to the current negative user feedback. \\

        \textbf{Error Propagation}
        & Inherits and further propagates an upstream error without being its original source. \\

        \textbf{Error Manifestation}
        & Exposes the erroneous content or result targeted by the current negative user feedback without necessarily being its source. \\

        \textbf{Recovered Error}
        & Previously introduced an error that was corrected or superseded before the current negative user feedback. \\

        \textbf{Supporting Context}
        & Provides information relevant to the current feedback issue without introducing or propagating the corresponding error. \\

        \textbf{Irrelevant}
        & Has no direct relation to the issue targeted by the current negative user feedback. \\

        \textbf{Missing Action}
        & A necessary action that the agent failed to perform, potentially causing the current negative user feedback. \\
        \bottomrule
    \end{tabularx}
    \caption{Definitions of candidate step roles.}
    \label{tab:stage3-role-definitions}
\end{table}

Furthermore, $\mathcal J_{\mathrm{step}}$ quantitatively assesses the relationship between $S_j$ and $F_i$ along multiple dimensions and outputs the structured assessment $A_{ij}$:
\begin{equation}
A_{ij}=\left(\ell_{ij},\mu_{ij},a_{ij},h_{ij},r_{ij},c_{ij}\right).
\label{eq:stage3-step-assessment}
\end{equation}
Here, $a_{ij}$ and $h_{ij}$ denote the error introduction and error inheritance scores of $S_j$, respectively; $r_{ij}$ denotes the relevance of $S_j$ to feedback $F_i$; $\mu_{ij}$ indicates whether $S_j$ directly manifests the error identified by the user; and $c_{ij}$ indicates whether $A_{ij}$ is supported by evidence from the original trajectory. For an inferred missing action $z$, SIRCA outputs its structured assessment $A_{iz}=(r_{iz},a_{iz},c_{iz})$ for subsequent analysis. The normalization and validation details of $A_{ij}$ are provided in Appendix~\ref{app:candidate-step-structured-assessment}.

\subsubsection{Interstep Information Propagation Modeling}

Errors identified by users may propagate across multiple steps; therefore, modeling information transfer relationships between steps is essential for accurate root cause localization. For the $i$th negative feedback record $F_i$, SIRCA classifies the relationship between a predecessor step $S_p$ and a successor step $S_j$ (where $p<j$) into two categories: information dependency and error propagation. The former indicates that step $S_j$ uses information produced by step $S_p$, whereas the latter indicates that an error relevant to negative feedback $F_i$ propagates from step $S_p$ to step $S_j$. The strengths of these two relationships are denoted by $\delta_{ipj}$ and $\eta_{ipj}$, respectively. Detailed computations are provided in Appendix~\ref{app:dependency-inference}.

\subsubsection{Negative Feedback Attribution Graph Construction}

After identifying the root cause candidate steps and their relationships, SIRCA constructs a negative feedback attribution graph $G=(\mathcal{V},\mathcal{L})$ to analyze how potential root causes lead to the observed negative user feedback. The graph is constructed as follows:

\paragraph{Node and Edge Construction.}
SIRCA creates nodes for each candidate step $S_j$, missing action hypothesis $z$, negative feedback item $F_i$, and related artifact, where the node corresponding to $F_i$ is denoted by $v_i^F$. Based on the analyzed relationships between trajectory steps, SIRCA establishes information dependency edges and error propagation edges between candidate steps, thereby connecting candidate steps to downstream steps that may be affected by them. In addition, to supplement potentially missing error propagation relations, SIRCA adds supplementary error propagation edges to the attribution graph based on LLM-based inference results and file interactions between steps. Detailed computations are provided in Appendix~\ref{app:supplementary-edge-weights}.
\paragraph{Edge Weight Computation.}
For an ordinary error propagation edge, its weight is computed as
\begin{equation}
w(S_p\!\rightarrow S_j)
=
\delta_{ipj}\eta_{ipj}\kappa_{ipj}.
\end{equation}

For the edge between an error manifestation step $S_j$ and the negative feedback node $v_i^F$, its weight is computed as
\begin{equation}
\begin{aligned}
w(S_j\!\rightarrow v_i^F)
=
\max\!\Bigl\{0.2,\,
&\max\!\left(r_{ij},0.55\mathbb{I}[j\in M_i^*]\right)\\
&\cdot\max(0.4,c_{ij})\Bigr\}.
\end{aligned}
\end{equation}

For the edge from a missing action hypothesis $z$ to an affected step $S_j$, its weight is computed as
\begin{equation}
w(z\!\rightarrow S_j)
=
r_{iz}a_{iz}\max(0.4,c_{iz}).
\end{equation}

Here, $\mathbb{I}[\cdot]$ is the indicator function; $\kappa_{ipj}=1$ if the interstep relationship is supported by evidence traceable to the trajectory, and $\kappa_{ipj}=0.4$ otherwise. For supplementary error propagation edges, the detailed computation of their weights is provided in Appendix~\ref{app:supplementary-edge-weights}.

Furthermore, SIRCA defines the overall path strength $P_i^{\mathrm{graph}}(v)$ from candidate node $v$ to negative feedback node $v_i^F$ as
\begin{equation}
P_i^{\mathrm{graph}}(v)
=
\max_{\pi\in\Pi_i(v)}
\left(\prod_{e\in\pi}w(e)\right)^{1/|\pi|}
\end{equation}
Here, $\Pi_i(v)$ contains paths from $v$ to $v_i^F$ consisting of positive weight edges, and $w(e)$ denotes edge $e$'s weight.

\paragraph{Weak Path Construction.}
For candidate steps lacking valid propagation paths, SIRCA introduces a weak path mechanism to prevent potential root causes from being excluded solely due to missing graph connections. When $P_i^{\mathrm{graph}}(S_j)=0$ and $S_j$ is not a pure error manifestation step with an error introduction score $a_{ij}<0.2$, SIRCA assigns a path strength of $P_i(S_j)=0.12$ for subsequent responsibility scoring if its role is Error Origin or Error Propagation, or if its relevance to the negative feedback satisfies $r_{ij}\geq0.4$. Such steps are then marked as weak path candidates to distinguish them from candidates with valid graph propagation paths.

\subsubsection{Root Cause Responsibility Scoring}


After obtaining the graph path strength of each candidate step, SIRCA uses the root cause responsibility score $\operatorname{Resp}_{ij}$ to jointly consider the error introduction and inheritance of each candidate step, as well as its propagation relationship with the current negative feedback, thereby quantifying each candidate step's root cause responsibility for the error indicated by the current feedback. The computation of $\operatorname{Resp}_{ij}$ is given in Eq.~\ref{eq:stage3-responsibility}.
\begin{equation}
\operatorname{Resp}_{ij}
=
\min\left\{
1,\,
c_{ij}\max(O_{ij},0.05)P_i(S_j)
\right\}.
\label{eq:stage3-responsibility}
\end{equation}

Here, $O_{ij}$ denotes the error origin strength of step $S_j$, and $P_i(S_j)$ denotes its path strength, which is usually set to the graph path strength $P_i^{\mathrm{graph}}(S_j)$. The computation of $O_{ij}$ is detailed in Appendix~\ref{app:Oij compute}.

\subsection{Static Intervention Validation}
Building on the results of the preceding stages, SIRCA constructs and evaluates static intervention proposals for eligible candidate objects and uses these assessments to determine the final root cause set $\mathcal C_i^*$.

\subsubsection{Static Intervention Eligibility Review}
\label{sec:Static Intervention Eligibility Review}
Not all root cause candidate steps are eligible for static intervention. SIRCA excludes steps in the candidate set $\mathcal R_i^{-}$ that are labeled as Irrelevant and simultaneously satisfy $r_{ij}<0.3$ and $a_{ij}<0.2$. Furthermore, among the retained candidate steps, a step is considered eligible for static intervention only if its result has not been overwritten or repaired by subsequent steps and its path strength satisfies $P_i(S_j)>0$. Let the steps eligible for intervention form the eligible root cause candidate set $\mathcal H_i$ for negative feedback $F_i$.

SIRCA then ranks the candidates in $\mathcal H_i$ by their root cause responsibility scores $\operatorname{Resp}_i(v)$ for subsequent static intervention. The ranking details are provided in Appendix~\ref{app:candidate-eligibility-ranking}.
\subsubsection{Static Intervention Construction and Evaluation}

For each negative feedback item $F_i$, SIRCA constructs static intervention proposals that could resolve the reported issue based on the eligible root cause candidate set $\mathcal H_i$. Specifically, SIRCA provides up to five top ranked candidates in $\mathcal H_i$ to the static intervention model $\mathcal J_{\mathrm{int}}$. The model $\mathcal J_{\mathrm{int}}$ proposes up to four hypothetical repairs, each of which may involve one or more candidate steps. Let the $r$th static intervention proposal $I_{ir}$ be represented as
\begin{equation}
I_{ir}
=
\left(
Z_{ir},
\Delta_{ir},
\operatorname{Suf}_{ir},
\operatorname{Min}_{ir},
E_{ir}
\right).
\end{equation}
Here, $Z_{ir}\subseteq\mathcal H_i$ denotes the set of candidate objects involved in intervention proposal $I_{ir}$, and $\Delta_{ir}$ denotes the corresponding hypothetical repair operation. $\operatorname{Suf}_{ir}$ denotes repair sufficiency and assesses the extent to which $\Delta_{ir}$ resolves the issue identified by $F_i$. $\operatorname{Min}_{ir}$ denotes repair minimality and assesses whether $\Delta_{ir}$ makes as few changes to the original trajectory as possible. $E_{ir}$ denotes the trajectory evidence supporting the intervention proposal. All valid static intervention proposals together form the intervention set $\mathcal I_i$.
\paragraph{Intervention Effectiveness Evaluation.} Applying a single intervention proposal may not block multiple error propagation paths associated with the same negative feedback. Therefore, SIRCA applies multiple intervention proposals jointly. Specifically, SIRCA enumerates all combinations of one to three intervention proposals from $\mathcal I_i$ and evaluates their repair effects. Let $\mathcal U\subseteq\mathcal I_i$ denote an intervention combination, and let $Z(\mathcal U)$ denote the set of candidate objects involved in its constituent proposals.

For each $\mathcal U$, SIRCA removes the nodes corresponding to $Z(\mathcal U)$ and their incident edges from the attribution graph $G$ to approximate its effect on blocking error propagation paths without replaying the trajectory. SIRCA then uses the final attribution score $\operatorname{Score}_{\mathrm{FA},i}(\mathcal U)$ to comprehensively evaluate its expected repair effect. The score is computed as follows:
\begin{equation}
\begin{aligned}
\operatorname{Score}_{\mathrm{FA},i}(\mathcal U)
={}&0.65\operatorname{Score}_{\mathrm{CF},i}(\mathcal U)\\
&+0.35\operatorname{Resp}_i^{\mathrm{rel}}(\mathcal U).
\end{aligned}
\end{equation}
Here, $\operatorname{Resp}_i^{\mathrm{rel}}(\mathcal U)$ denotes the relative responsibility strength of combination $\mathcal U$, whose computation is detailed in Appendix~\ref{app:intervention-combination-scoring}. The term $\operatorname{Score}_{\mathrm{CF},i}(\mathcal U)$ denotes the static counterfactual repair score of $\mathcal U$, computed as
\begin{equation}
\begin{aligned}
\operatorname{Score}_{\mathrm{CF},i}(\mathcal U)
={}&0.50B_i\!\left(Z(\mathcal U)\right)
+0.35\operatorname{Suf}_i(\mathcal U)\\
&+0.15\operatorname{Min}_i(\mathcal U)
-\operatorname{Pen}(\mathcal U).
\end{aligned}
\end{equation}
Here, $B_i(Z(\mathcal U))$ denotes the path blocking rate, reflecting the change in the total graph path strength before and after the intervention; $\operatorname{Suf}_i(\mathcal U)$ and $\operatorname{Min}_i(\mathcal U)$ denote the repair sufficiency and repair minimality of $\mathcal U$, respectively. The detailed computations of these components are also provided in Appendix~\ref{app:intervention-combination-scoring}. The term $\operatorname{Pen}(\mathcal U)=0.03(|\mathcal U|-1)$ is a combination size penalty, making SIRCA favor combinations with fewer intervention proposals when their repair effects are comparable. Note that both \(\operatorname{Score}_{\mathrm{FA},i}(\mathcal U)\) and \(\operatorname{Score}_{\mathrm{CF},i}(\mathcal U)\) are constrained to the range \([0,1]\).

\subsubsection{Final Root Cause Set Selection}

SIRCA considers an intervention combination $\mathcal U$ feasible when it satisfies both of the following conditions:
\begin{equation}
B_i\!\left(Z(\mathcal U)\right)\geq 0.8,
\qquad
\operatorname{Suf}_i(\mathcal U)\geq 0.65.
\label{eq:stage4-feasible-set}
\end{equation}
Among feasible combinations, SIRCA first compares their coverage of the top five eligible candidates ranked by root cause responsibility score $\operatorname{Resp}_i(v)$. When coverage is equal, SIRCA prefers combinations with fewer intervention proposals, higher final attribution scores $\operatorname{Score}_{\mathrm{FA},i}(\mathcal U)$, shorter distances to the error manifestation step, and higher static counterfactual repair scores $\operatorname{Score}_{\mathrm{CF},i}(\mathcal U)$, in that order.

Let $\mathcal U_i^*$ denote the combination ranked first under the rules. Its candidate objects form the initial root cause set $\mathcal C_i^{(0)}=Z(\mathcal U_i^*)$. SIRCA attempts to add adjacent candidates, uncovered priority candidates, and earlier candidate steps to $\mathcal C_i^{(0)}$. SIRCA accepts a correction only if additions pass independent validation, improve the final attribution score $\operatorname{Score}_{\mathrm{FA},i}$, and do not render an originally feasible combination infeasible. This yields the final root cause set $\mathcal C_i^*$.

\section{Results}
\label{sec:results}


\subsection{Experimental Setup}
\label{sec:experimental-setup}

\paragraph{Datasets.}We evaluate SIRCA's root cause localization performance on our wireless agent dataset and the public Who\&When trajectory dataset~\cite{zhang2025agent}. The wireless agent dataset contains 50 samples with explicit negative user feedback and root cause annotations, collected from telecommunications network operations and wireless network optimization scenarios. For each trajectory, we annotate one or more root cause steps associated with the negative user feedback. Who\&When contains 126 relatively short Algorithm Generated (AG) trajectories and 58 longer Hand Crafted (HC) trajectories, each annotated with one decisive error step. Since the dataset requires identifying a single root cause step, we adapt SIRCA's output accordingly, as detailed in Appendix~\ref{app:public-trajectory-adaptation}.

\paragraph{Compared Methods and Evaluation Metrics.}We select baselines and evaluation metrics according to the task setting of each dataset. For the Who\&When dataset, we use All at Once (AAO)~\cite{zhang2025agent}, Step by Step (SBS)~\cite{zhang2025agent}, Binary Search (BS)~\cite{zhang2025agent}, and CHIEF~\cite{wang2026flat} as baselines. We evaluate localization performance using Exact Root Cause Localization Accuracy $\mathrm{Acc}_{\mathrm{step}}^{\mathrm{exact}}$ and Tolerance Aware Root Cause Localization Accuracy $\mathrm{Acc}_{\mathrm{step}}^{(\pm1)}$.
For the wireless agent dataset constructed in this work, existing public methods are not directly applicable to localization of multiple root causes based on explicit negative user feedback. We therefore adapt SBS and AAO, which have broader applicability, as baselines for this dataset. We evaluate localization performance using trajectory averaged root cause precision $\mathrm{Traj\text{-}P}$, recall $\mathrm{Traj\text{-}R}$, and $F_1$ score $\mathrm{Traj\text{-}F}_1$, as well as the any root cause step hit rate $\mathrm{Hit}_{\mathrm{step}}^{\mathrm{any}}$. The computation of all evaluation metrics is detailed in Appendix~\ref{app:evaluation-metrics}. 

In addition, to keep the model configuration consistent across task settings, all LLMs responsible for judgment and inference in SIRCA use DeepSeek-V4-Flash. For each dataset, all methods are run independently three times under the same experimental settings.

\subsection{Results on the Who\&When Dataset}
\label{sec:results-whowhen}

\begin{figure}[h]
    \centering
    \includegraphics[width=\linewidth]{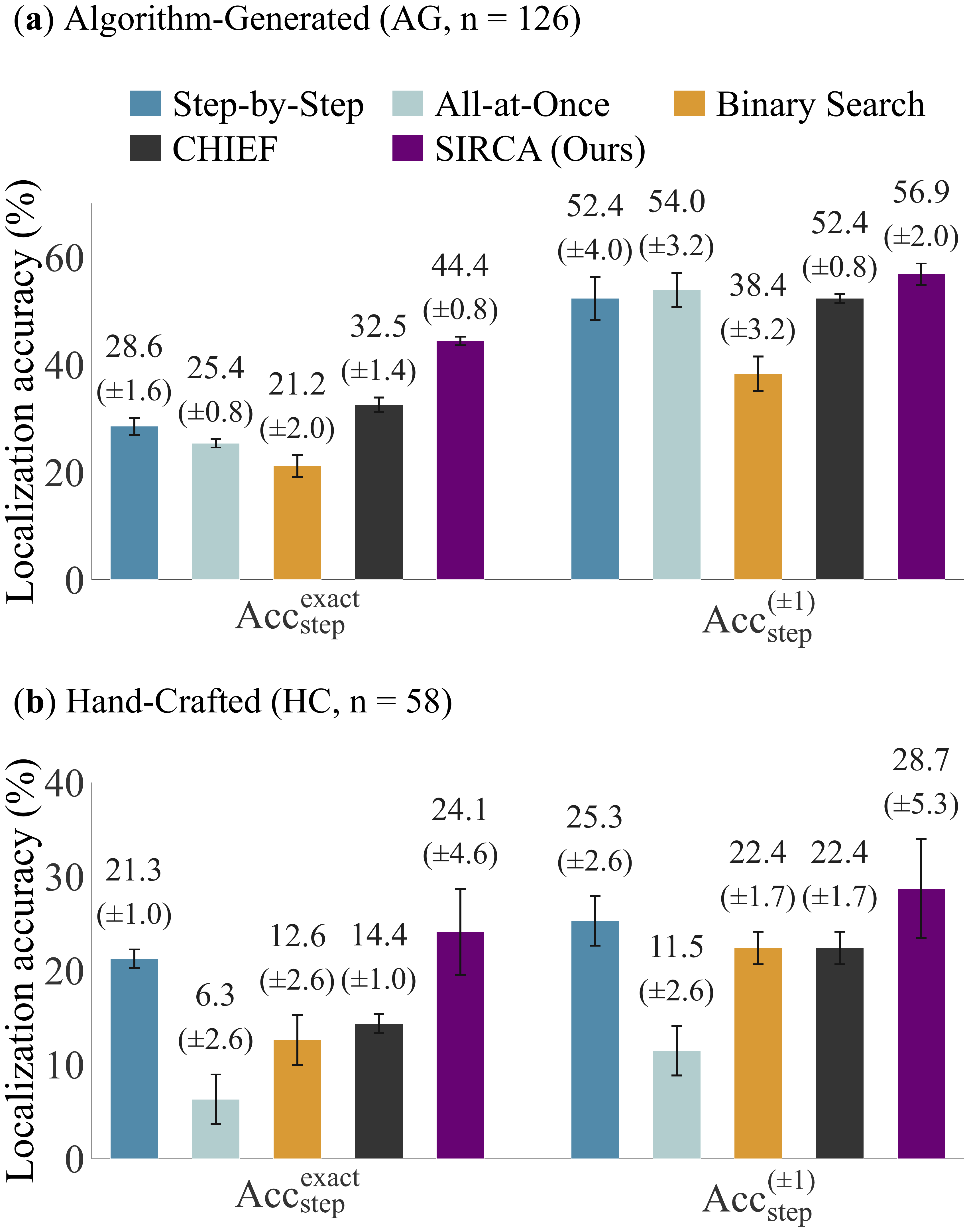}
    
    \caption{Root cause localization accuracy of different methods on AG and HC subsets of Who\&When.}
    \label{fig:ag result}
\end{figure}

As shown in Fig.~\ref{fig:ag result}, on the AG dataset, SIRCA achieves an Exact Root Cause Localization Accuracy $\mathrm{Acc}_{\mathrm{step}}^{\mathrm{exact}}$ of 44.44\% and a Tolerance Aware Root Cause Localization Accuracy $\mathrm{Acc}_{\mathrm{step}}^{(\pm1)}$ of 56.88\%, representing relative improvements of 36.59\% and 5.39\% over CHIEF and AAO, the best performing baselines on the respective metrics. 

On the long trajectory HC dataset, both localization accuracy and stability decline for all methods compared with AG, indicating that error propagation across steps in long trajectories poses a greater challenge for root cause localization. Nevertheless, SIRCA achieves the best results on both $\mathrm{Acc}_{\mathrm{step}}^{\mathrm{exact}}$ and $\mathrm{Acc}_{\mathrm{step}}^{(\pm1)}$, with relative improvements of 13.51\% and 13.64\%, respectively, over SBS, the best performing baseline on both metrics.

Further analysis shows that, when the permitted localization error is relaxed, the localization accuracy of the baseline methods increases by an average of 22.35 and 6.75 percentage points on AG and HC, respectively, whereas SIRCA improves by only 12.43 and 4.60 percentage points. This indicates that, although the baseline methods can locate steps near the decisive error, they are more likely to confuse the step that first introduces the error with adjacent steps that propagate or manifest it. In contrast, by jointly modeling error origin and propagation, SIRCA can identify root cause boundaries more accurately. 

\begin{table*}[t]
    \centering
    \small
    \setlength{\tabcolsep}{4pt}
    \renewcommand{\arraystretch}{1.15}

    \begin{tabular*}{\textwidth}{@{\extracolsep{\fill}}lcccccccc@{}}
    \toprule
    \multirow{2}{*}{\textbf{Model}}
      & \multicolumn{2}{c}{\textbf{Algorithm Generated}}
      & \multicolumn{2}{c}{\textbf{Hand Crafted}}
      & \multicolumn{4}{c}{\textbf{Wireless Agent Dataset}} \\
    \cmidrule(lr){2-3}
    \cmidrule(lr){4-5}
    \cmidrule(lr){6-9}
      & \shortstack{$\mathrm{Acc}_{\mathrm{step}}^{\mathrm{exact}}$}
      & \shortstack{$\mathrm{Acc}_{\mathrm{step}}^{(\pm 1)}$}
      & \shortstack{$\mathrm{Acc}_{\mathrm{step}}^{\mathrm{exact}}$}
      & \shortstack{$\mathrm{Acc}_{\mathrm{step}}^{(\pm 1)}$}
      & $\mathrm{Traj}\text{-}F_1$
      & \shortstack{$\mathrm{Hit}_{\mathrm{step}}^{\mathrm{any}}$}
      & $\mathrm{Traj}\text{-}R$
      & $\mathrm{Traj}\text{-}P$ \\
    \midrule
    Kimi-K3 & 39.68 & 52.38 & \textbf{31.03} & 36.21 & 20.57 & 40.00 & 30.33 & 18.13 \\
    GPT-5.6-sol & 42.06 & 57.94 & 27.59 & 39.66 & 21.05 & 44.00 & 33.12 & 20.23 \\
    Grok-4.6 & 42.86 & 57.14 & 27.59 & \textbf{41.38} & 23.10 & 46.00 & 34.82 & \textbf{20.37} \\
    Gemini-3.8-Flash & \textbf{51.59} & \textbf{61.11} & 25.86 & 36.21 & 21.38 & \textbf{50.00} & 33.71 & 19.69 \\
    DeepSeek-V4-Flash & 43.65 & 58.73 & 22.41 & 27.59 & \textbf{23.64} & \textbf{50.00} & \textbf{37.55} & 20.07 \\
    \bottomrule
    \end{tabular*}

    \caption{Root cause localization performance of SIRCA across backbone models (\%).}
    \label{tab:backbone-ablation-cn}
\end{table*}
\subsection{Results on the Wireless Agent Dataset}
\label{sec:results-domain}

\begin{figure}[h]
    \centering
    \includegraphics[width=\linewidth]{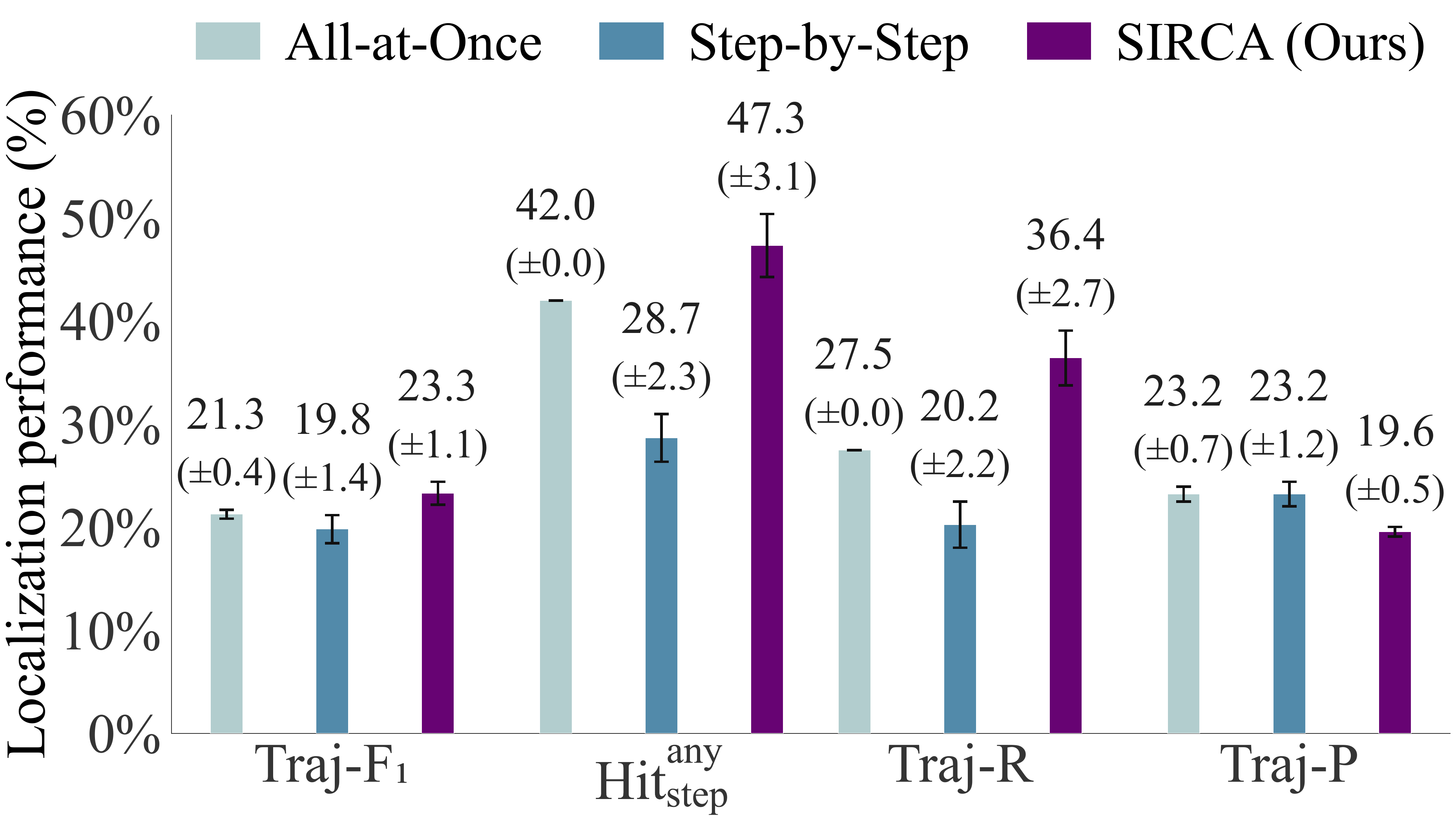}
    
    \caption{Comparison of root cause localization performance across different methods on the multi root cause wireless agent dataset.}
    \label{fig:domain result}
\end{figure}

Unlike Who\&When, the wireless agent dataset constructed in this work evaluates whether different methods can leverage negative user feedback to localize multiple potential root causes. Fig.~\ref{fig:domain result} presents the root cause localization results of different methods on this dataset.

As shown in Fig.~\ref{fig:domain result}, SIRCA achieves a trajectory averaged $F_1$ score $\mathrm{Traj\text{-}F}_1$ of $23.30\%$, an any root cause step hit rate $\mathrm{Hit}_{\mathrm{step}}^{\mathrm{any}}$ of $47.33\%$, and a trajectory averaged recall $\mathrm{Traj\text{-}R}$ of $36.43\%$. Compared with AAO, the strongest baseline on all three metrics, these results correspond to relative improvements of $9.58\%$, $12.70\%$, and $32.52\%$, respectively. This indicates that SIRCA can identify root causes more effectively and cover more ground truth root cause steps. However, its trajectory averaged precision $\mathrm{Traj\text{-}P}$ is $19.56\%$, lower than the $23.21\%$ achieved by AAO and the $23.22\%$ achieved by SBS. This suggests that SIRCA's relatively permissive candidate retention strategy reduces missed root causes, but may also retain steps that are related to the negative feedback without being actual root causes.

\subsection{Sensitivity Analysis of Backbone LLMs}
\label{app:Performance Comparison across Different
Backbone LLMs}




Since SIRCA's root cause localization capability relies heavily on large language models, this section further evaluates its performance across different backbone models, with the results presented in Table~\ref{tab:backbone-ablation-cn}.

The results show that SIRCA exhibits a certain degree of cross model applicability. Across different backbone models, its $\mathrm{Acc}_{\mathrm{step}}^{\mathrm{exact}}$ is 43.97\% on AG and 26.90\% on HC, respectively, while its $\mathrm{Traj\text{-}F}_1$ on the wireless agent dataset is 21.95\%. The overall results are broadly consistent with those reported in Sections~\ref{sec:results-whowhen} and ~\ref{sec:results-domain}. Notably, no clear positive correlation is observed between the overall capabilities of backbone models and root cause localization accuracy, indicating that more powerful backbone models do not necessarily lead to better localization performance.

\section{Conclusion}

This paper investigates the problem of root cause localization in agent trajectories based on negative user feedback. Errors identified by user feedback may propagate across multiple interdependent trajectory steps, making their origins difficult to identify accurately. To address this challenge, we propose SIRCA, a root cause localization method that does not require replaying agent execution. SIRCA combines LLM reasoning with programmatic analysis, establishes attribution anchors from negative user feedback, constructs alternative anchors from task failure information when explicit negative feedback is unavailable, and models step dependencies and error propagation through an attribution graph. Finally, SIRCA determines the final root causes through LLM-based evaluation of static intervention repair proposals.

On the AG and HC subsets of the Who\&When dataset, SIRCA achieves relative improvements of $36.59\%$ and $13.51\%$, respectively, in exact root cause localization accuracy over the strongest baseline on each subset. In addition, we construct a domain specific wireless agent trajectory dataset containing 50 trajectories with explicit negative user feedback and root cause step annotations. On this dataset, SIRCA achieves relative improvements of $9.58\%$ and $32.52\%$ in trajectory averaged $F_1$ and recall, respectively, over the strongest baseline. We also publicly release the dataset and SIRCA source code to support future research.

\section*{Limitations}

This work has the following main limitations:

\paragraph{Root cause analysis relies on the judgment capabilities of large language models.}
    SIRCA relies on large language models for step role assessment, relation inference, and repair effect evaluation. Errors in their judgments may affect the final attribution results. Therefore, future work could combine cross validation across multiple models with stricter trajectory evidence constraints to improve the reliability of attribution judgments.

\paragraph{Localization performance on complex long trajectories remains open to improvement.}
    SIRCA achieves lower localization accuracy on HC than on AG, indicating that root cause localization in complex long trajectories remains challenging. Future work could explore hierarchical trajectory modeling and dependency tracking across steps to improve the identification of distant error origins and their propagation.

\appendix

\section{Validation of Negative Feedback Record $\hat{F}_i$}
\label{sec:Negative-Feedback Validation}

\begin{figure*}[t]
    \centering
    \includegraphics[width=\linewidth]{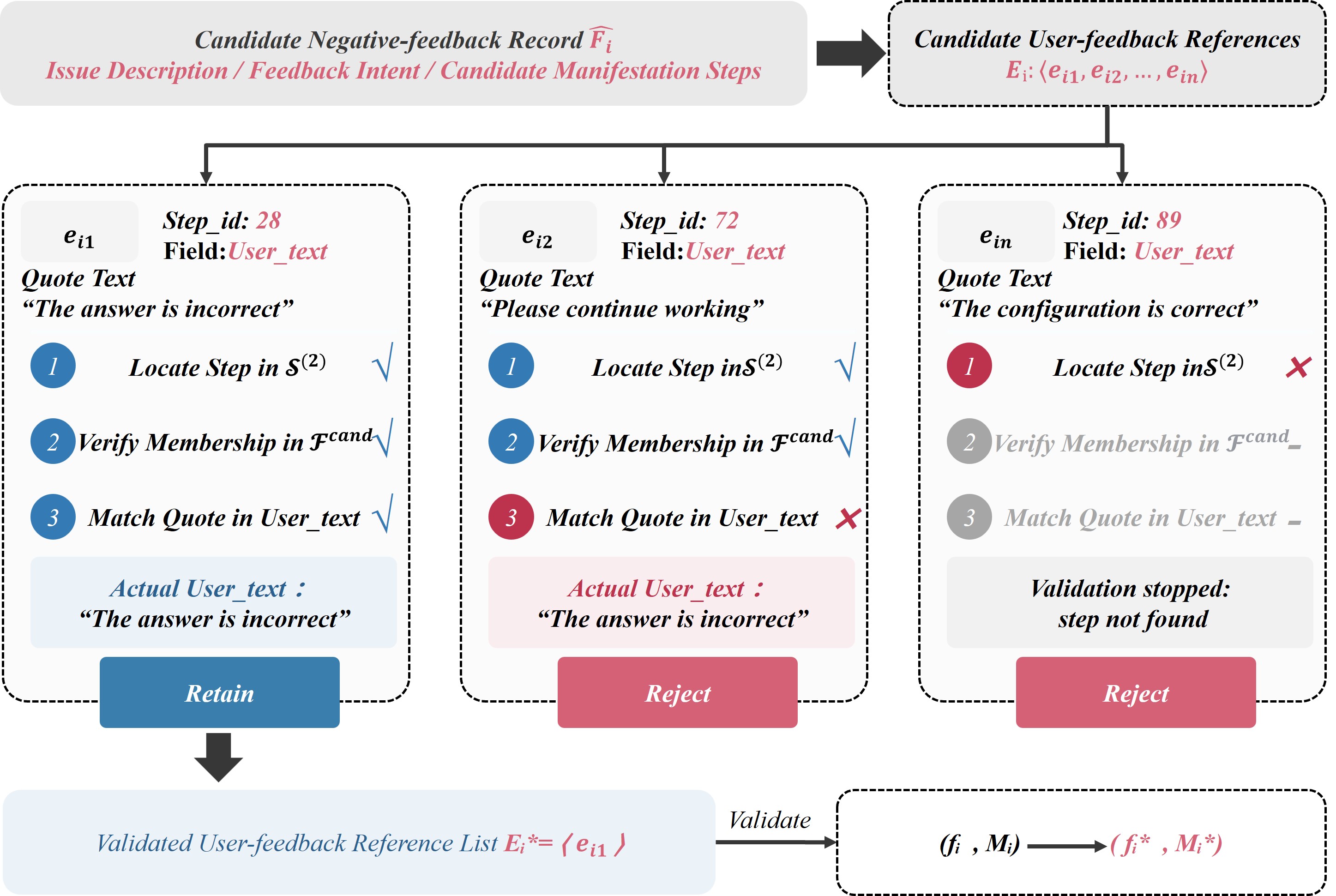}
    
    \caption{Validation procedure for candidate negative feedback steps}
    \label{fig:Validation procedure for candidate negative-feedback steps}
\end{figure*}

Since the candidate negative feedback record $\hat{F}_i$ returned by the feedback classification model $\mathcal{J}$ may contain content unsupported by the actual trajectory, it cannot be directly used for subsequent attribution analysis. Therefore, SIRCA validates the candidate textual references $E_i$ against the structured trajectory context $\mathcal{S}^{(2)}$ to determine whether the candidate negative feedback record $\hat{F}_i$ is supported by the actual trajectory. The validation procedure is illustrated in Figure~\ref{fig:Validation procedure for candidate negative-feedback steps}.

Specifically, for each candidate negative feedback record $\hat{F}_i$, the feedback classification model $\mathcal{J}$ returns a list of candidate textual references:
\begin{equation}
    E_i
=
\left\langle
e_{i1},e_{i2},\ldots,e_{in}
\right\rangle .
\end{equation}


Each candidate textual reference $e$ contains three main fields: \texttt{step\_id} identifies the structured trajectory step containing the referenced content; \texttt{field} specifies the text field being referenced within that step, such as \texttt{user\_text}, which represents the user feedback text; and \texttt{quote} stores the specific text fragment selected by the feedback classification model $\mathcal{J}$ from the field specified by \texttt{field} in the step identified by \texttt{step\_id}.

For each candidate textual reference $e$, SIRCA sequentially performs the following validation checks:

\begin{itemize}
    \item SIRCA locates the corresponding step in $\mathcal{S}^{(2)}$ according to \texttt{step\_id}. If the step does not exist, the reference is considered invalid.
    \item If the step is successfully located, SIRCA checks whether it belongs to the candidate feedback step set $\mathcal{F}^{\mathrm{cand}}$.
    \item SIRCA checks whether \texttt{quote} can be matched against the actual text in the corresponding \texttt{field}.
\end{itemize}

A candidate textual reference is considered valid only if all the above conditions are satisfied. All valid textual references are retained in their original order to form the validated textual reference list $E_i^*$. To improve the accuracy of the textual references, SIRCA replaces the \texttt{quote} in each candidate reference with the original text fragment actually matched in the corresponding \texttt{field}, thereby ensuring that every textual reference in $E_i^*$ can be traced back to the original trajectory. A candidate negative feedback record must contain at least one valid textual reference pointing to the corresponding user feedback step before it can be used in subsequent attribution analysis.


After obtaining $E_i^*$, SIRCA further validates the negative feedback step index $f_i$ generated by the feedback classification model $\mathcal{J}$, yielding the validated step index $f_i^*$. Subsequently, SIRCA uses $f_i^*$ as the temporal boundary to further validate the candidate manifestation step set $M_i$ and its corresponding agent content references. Only steps that occur before this boundary, are generated by the agent, and are supported by direct textual evidence are retained, yielding the validated manifestation step set $M_i^*$.

\section{Implementation Details of the Feedback Graph Construction}

\subsection{Candidate Step Role Assessment Examples}
\label{app:candidate-step-role-assessment}

This section illustrates, through examples, how the step assessment model $\mathcal{J}_{\mathrm{step}}$ determines the role of each candidate step. Specifically, $\mathcal{J}_{\mathrm{step}}$ considers the negative feedback query $u_i$, together with the action summary, action input, observation summary, observation content, and candidate textual evidence associated with step $S_j$, to determine the role of the candidate step with respect to the negative feedback record $F_i$. Representative assessment examples are provided in Table~\ref{tab:stage3-role-examples}.

\begin{table*}[t]
    \centering

    \small
    \setlength{\tabcolsep}{6pt}
    \renewcommand{\arraystretch}{1.15}
    \renewcommand{\tabularxcolumn}[1]{m{#1}}

    \begin{tabularx}{\textwidth}{
        @{}
        >{\raggedright\arraybackslash}m{0.08\textwidth}
        >{\raggedright\arraybackslash}X
        >{\raggedright\arraybackslash}m{0.38\textwidth}
        @{}
    }
        \toprule
        \textbf{Step}
        & \textbf{Trajectory Content}
        & \textbf{Assessment Result} \\
        \midrule

        $S_5$
        & Agent: ``There is no need to inspect the configuration; directly assume that the parameter is 40.''
        & \textbf{Error Origin}: First introduces the incorrect parameter that ultimately leads to the current user negative feedback. \\[2pt]

        $S_{12}$
        & Agent: ``Continue the calculation using the previously determined parameter value of 40.''
        & \textbf{Error Propagation}: Inherits and further propagates the incorrect parameter introduced by an upstream step. \\[2pt]

        $S_{15}$
        & Agent: ``The current network connection is operating normally.''
        & \textbf{Irrelevant}: Has no direct relation to the incorrect parameter value identified by the current user negative feedback. \\[2pt]

        $S_{20}$
        & The agent directly presents an incorrect calculation result to the user.
        & \textbf{Error Manifestation}: Exposes the erroneous result targeted by the current feedback without being its original source. \\[2pt]

        $S_{21}$
        & User: ``The configuration specifies 42. Why did you use 40?''
        & \textbf{User Negative Feedback}. \\

        \bottomrule
    \end{tabularx}
    \caption{Examples of candidate step role assessment.}
    \label{tab:stage3-role-examples}
\end{table*}

\subsection{Normalization and Validation of Candidate Step Assessment $A_{ij}$}

\label{app:candidate-step-structured-assessment}
\setcounter{validationstep}{0}

This section describes how SIRCA validates and normalizes the raw assessment generated by the step assessment model $\mathcal{J}_{\mathrm{step}}$. 

For a negative feedback record $F_i$ and a candidate step $S_j$, the step assessment model first generates a raw assessment, which is represented as:

\begin{equation}
\mathcal{J}_{\mathrm{step}}\!\left(F_i,S_j\right)
=
\left(
\hat{\ell}_{ij},
\mu_{ij},
\hat{a}_{ij},
\hat{h}_{ij},
\hat{r}_{ij},
\widehat{\mathcal{E}}_{ij}
\right)
\end{equation}
Here, $\hat{\ell}_{ij}$ is the role label, and $\mu_{ij}$ indicates whether candidate step $S_j$ directly manifests the error identified in the user's negative feedback. The indicator is assigned a value of 1 if the step directly presents an erroneous result to the user or explicitly manifests the reported issue, and 0 otherwise. The values $\hat{a}_{ij}$, $\hat{h}_{ij}$, and $\hat{r}_{ij}$ are the raw error introduction, error inheritance, and feedback relevance scores, respectively. $\widehat{\mathcal{E}}_{ij}$ denotes the set of candidate textual evidence references.

SIRCA then validates and normalizes the raw assessment through the following stages, thereby obtaining the final structured assessment $A_{ij}$:

\stepheading{Assessment Validity Verification}

To prevent $\mathcal{J}_{\mathrm{step}}$ from generating assessments unsupported by the actual trajectory, SIRCA uses the candidate textual evidence references $\widehat{\mathcal{E}}_{ij}$ provided by the model to check whether these references can be matched to the actual content of candidate step $S_j$. Except for steps labeled as Irrelevant, each role assessment must contain at least one valid textual evidence reference; otherwise, the assessment is considered invalid. If at least one valid textual evidence reference exists, the evidence support indicator is set to $c_{ij}=1$; otherwise, $c_{ij}=0$.

\stepheading{Assessment Score Normalization}

SIRCA applies a normalization function $Q(\cdot)$ to constrain model generated scores to $[0,1]$ and round them to one decimal place:
\begin{equation}
Q(x)
=
\frac{1}{10}
\left\lfloor
10\operatorname{clip}(x,0,1)+\frac{1}{2}
\right\rfloor
\end{equation}
where
\begin{equation}
\operatorname{clip}(x,0,1)
=
\min\{\max\{x,0\},1\}.
\end{equation}
The normalized scores are
\begin{equation}
\bar{a}_{ij}=Q(\hat{a}_{ij}),\quad
\bar{h}_{ij}=Q(\hat{h}_{ij}),\quad
\bar{r}_{ij}=Q(\hat{r}_{ij}).
\end{equation}

\stepheading{Role Consistency Correction}

After evidence validation and score normalization, SIRCA corrects the role label according to the rules in Table~\ref{tab:stage3-role-correction}.

\begin{table*}[t]
    \centering
    \small
    \setlength{\tabcolsep}{6pt}
    \renewcommand{\arraystretch}{1.25}

    \begin{tabularx}{\textwidth}{
        @{}
        >{\raggedright\arraybackslash}X
        >{\raggedright\arraybackslash}m{0.22\textwidth}
        @{}
    }
        \toprule
        \textbf{Condition} & \textbf{Corrected Role} \\
        \midrule

        The step has an error manifestation indicator.
        & \textbf{Error Manifestation} \\

        The step is labeled as an error origin, but its error inheritance score is at least $0.6$.
        & \textbf{Error Propagation} \\

        A step labeled as supporting context or irrelevant is not marked as an error manifestation but shares a numerical value or filename with the feedback issue.
        & \textbf{Error Propagation} \\

        After the preceding corrections, the step remains irrelevant, is not marked as an error manifestation, and has a feedback relevance score of at least $0.6$.
        & \textbf{Supporting Context} \\

        \bottomrule
    \end{tabularx}

    \caption{Role consistency correction rules.}
    \label{tab:stage3-role-correction}
\end{table*}

\stepheading{Score Consistency Correction}

Given the corrected role, SIRCA further adjusts the error introduction and error inheritance scores as shown in Table~\ref{tab:stage3-score-constraints}.

\begin{table*}[t]
    \centering
    \small
    \setlength{\tabcolsep}{6pt}
    \renewcommand{\arraystretch}{1.25}
    \renewcommand{\tabularxcolumn}[1]{m{#1}}

    \begin{tabularx}{\textwidth}{
        @{}
        >{\raggedright\arraybackslash}X
        >{\centering\arraybackslash}m{0.25\textwidth}
        >{\centering\arraybackslash}m{0.25\textwidth}
        @{}
    }
        \toprule
        \textbf{Corrected Role}
        &
        \shortstack[c]{\textbf{Error Introduction} $a_{ij}$}
        &
        \shortstack[c]{\textbf{Error Inheritance} $h_{ij}$}
        \\
        \midrule

        \textbf{Error Origin}
        & $\max(\bar{a}_{ij},0.6)$
        & $\min(\bar{h}_{ij},0.3)$ \\

        \textbf{Error Propagation}
        & $\min(\bar{a}_{ij},0.3)$
        & $\max(\bar{h}_{ij},0.6)$ \\

        \textbf{Error Manifestation}
        & $\min(\bar{a}_{ij},0.3)$
        & $\bar{h}_{ij}$ \\

        \textbf{Supporting Context} ($\bar{r}_{ij}\geq 0.6$)
        & $\min(\bar{a}_{ij},0.5)$
        & $\bar{h}_{ij}$ \\

        \textbf{Supporting Context} ($\bar{r}_{ij}<0.6$)
        & $\min(\bar{a}_{ij},0.3)$
        & $\bar{h}_{ij}$ \\

        \textbf{Irrelevant}
        & $\min(\bar{a}_{ij},0.2)$
        & $\min(\bar{h}_{ij},0.2)$ \\

        \textbf{Recovered Error}
        & $\bar{a}_{ij}$
        & $\bar{h}_{ij}$ \\

        \bottomrule
    \end{tabularx}

    \caption{Score consistency correction rules.}
    \label{tab:stage3-score-constraints}
\end{table*}

For a step corrected to error propagation because it shares a numerical value or filename with the feedback issue, SIRCA also imposes lower bounds on feedback relevance and error inheritance:
\begin{equation}
r_{ij}=\max(\bar{r}_{ij},0.4),\quad
h_{ij}=\max(\bar{h}_{ij},0.6).
\end{equation}
In all other cases, feedback relevance remains at its normalized value, $r_{ij}=\bar{r}_{ij}$. The resulting structured assessment is
\begin{equation}
A_{ij}
=
\left(
\ell_{ij},
\mu_{ij},
a_{ij},
h_{ij},
r_{ij},
c_{ij}
\right).
\end{equation}

\subsection{Computation of Interstep Dependency Strength $\delta_{ipj}$ and Error Propagation Strength $\eta_{ipj}$}
\label{app:dependency-inference}


SIRCA computes the information dependency strength $\delta_{ipj}$ and error propagation strength $\eta_{ipj}$ between predecessor step $S_p$ and successor step $S_j$ as follows:
\begin{subequations}
\label{eq:dependency-propagation-strength}
\begin{align}
\eta_{ipj}
&=\max\!\left\{
h_{ij},
a_{ip}\max\!\left(r_{ij},0.55\chi_{ij}^{\mathrm{man}}\right)
\right\},
\label{eq:error-propagation-strength}\\
\delta_{ipj}
&=\min\{1,g_{ipj}\},
\label{eq:dependency-strength}
\end{align}
\end{subequations}
Here, $\chi_{ij}^{\mathrm{man}}=1$ if $j\in M_i^*$,
$\mu_{ij}=1$, or $\ell_{ij}=\texttt{manifestation}$,
and is $0$ otherwise. The term $g_{ipj}$ denotes the directed association score from $S_p$ to
$S_j$ with respect to the $i$th negative feedback instance. It aggregates
multiple interstep association signals and is computed as
\begin{equation}
\label{eq:inter-step-association-score}
\begin{aligned}
g_{ipj}
={}&
0.30L_{ipj}
+0.25\min\left\{1,n^{\mathrm{art}}_{ipj}\right\}
\\
&+
0.15\min\left\{1,0.35n^{\mathrm{num}}_{ipj}\right\}
+\frac{0.10}{j-p}
\\
&+
\beta_{ipj}
+0.35w_{pj}
+0.25t_{pj}.
\end{aligned}
\end{equation}
Here, $L_{ipj}$ denotes the term overlap ratio between the textual evidence cited in the assessment results $A_{ip}$ and $A_{ij}$ of steps $S_p$ and $S_j$, respectively, with respect to negative feedback $F_i$.
The quantity $n^{\mathrm{art}}_{ipj}$ counts the file paths, file names,
and tool names shared by the two steps, while
$n^{\mathrm{num}}_{ipj}$ counts their shared distinct numerical values.
The term $1/(j-p)$ captures the temporal proximity between the two steps. The term $w_{pj}$ represents the file level read after write dependency
between $S_p$ and $S_j$. Specifically, $w_{pj}=1$ if $S_p$ writes to a
file that is subsequently read by $S_j$, and $w_{pj}=0$ otherwise.
The term $t_{pj}$ captures the tool call association between the two
steps. It is set to $1$ when the two steps share the same nonempty
\texttt{tool\_use\_id}, and to $0$ otherwise. The term $\beta_{ipj}$ denotes the role compatibility reward, which measures whether the roles of $S_p$ and $S_j$ are consistent with the pattern in which an error propagates from a preceding step to a
subsequent step. It is computed as follows.
\begin{equation}
\beta_{ipj}=
\begin{cases}
0.40,
& \chi_{ip}^{\mathrm{pre}}=1
  \land \chi_{ij}^{\mathrm{man}}=1,\\[2pt]
0.35,
& \chi_{ip}^{\mathrm{pre}}
  \chi_{ij}^{\mathrm{post}}
  (1-\chi_{ij}^{\mathrm{man}})=1,\\[2pt]
0,
& \text{otherwise}.
\end{cases}
\label{eq:role_compatibility_reward}
\end{equation}
Here, $\chi_{ik}^{\mathrm{pre}}$ and $\chi_{ik}^{\mathrm{post}}$ indicate whether step $S_k$ satisfies the upstream and downstream role conditions, respectively, for computing the role compatibility bonus.
$\chi_{ik}^{\mathrm{pre}}=1$ for error origins, error propagation steps, error manifestations, recovered errors, and supporting context steps with $r_{ik}\geq 0.4$.
$\chi_{ik}^{\mathrm{post}}=1$ for error propagation steps, error manifestations, and supporting context steps with $r_{ik}\geq 0.4$.
In all other cases, the corresponding indicator is set to $0$.


\subsection{Computation Details of Supplementary Error Propagation Edges}
\label{app:supplementary-edge-weights}

To address potentially missing interstep information propagation relationships in the negative feedback attribution graph, SIRCA further combines large language model analysis with file interaction information between steps to supplement potential error propagation edges and calculate their corresponding weights.

\subsubsection{LLM-Based Supplementation of Error Propagation Edges}

SIRCA invokes the LLM-based dependency inference model $\mathcal{J}_{\mathrm{dep}}$ and takes the current negative feedback record $F_i$, the structured assessments of candidate steps $\{A_{ik}\}_{k\in K_i}$, and the step pairs $(S_p,S_j)$ satisfying $p,j\in K_i$ and $p<j$ as input to identify potentially missed interstep dependency and error propagation relationships. Here, $K_i$ denotes the set of candidate step indices considered in the current relationship inference. For each inferred relationship $S_p\rightarrow S_j$, $\mathcal{J}_{\mathrm{dep}}$ outputs the dependency strength $\delta_{ipj}$, the error propagation strength $\eta_{ipj}$, and the corresponding candidate textual evidence references. SIRCA then checks whether both strengths are finite values within $[0,1]$ and verifies whether the candidate textual evidence can be grounded in the actual trajectory content of both the predecessor step $S_p$ and the successor step $S_j$. Only relationships that pass both the score checks and evidence validation are retained.

\subsubsection{Supplementing Error Propagation Edges with File Interactions}
For each negative feedback record $F_i$, SIRCA supplements potential error propagation edges between candidate steps that have been assessed and included in the attribution graph. Specifically, for a predecessor step $S_p$ and a successor step $S_j$, SIRCA adds an error propagation edge from $S_p$ to $S_j$ if at least one of the two steps has a feedback relevance score $r_{ij}\geq 0.25$ and at least one of the following file association conditions is satisfied:
\begin{itemize}
    \item $S_p$ and $S_j$ have a file write and read relationship.
    \item $S_p$ and $S_j$ share a file path.
    \item Matching file or path identifiers appear in the action summaries, action inputs, or observation summaries of the two steps.
\end{itemize}

\subsubsection{Weight Computation of Supplementary Error Propagation Edges}

For error propagation edges inferred by the LLM-based model $\mathcal{J}_{\mathrm{dep}}$, their weights in the attribution graph are computed in the same way as those of ordinary error propagation edges:
\begin{equation}
w(S_p\rightarrow S_j)
=\delta_{ipj}\eta_{ipj}\kappa_{ipj},
\end{equation}
where $\delta_{ipj}$ denotes the dependency strength and $\eta_{ipj}$ denotes the error propagation strength. If the relationship is supported by textual evidence that can be traced back to the original trajectory, $\kappa_{ipj}=1$; otherwise, $\kappa_{ipj}=0.4$.

For error propagation edges directly supplemented based on file interaction information, their weights are determined according to the file association types between steps, as shown in Table~\ref{tab:stage3-supplementary-edge-weights}.

\begin{table}[t]
    \centering
    \small
    \setlength{\tabcolsep}{4pt}
    \renewcommand{\arraystretch}{1.25}
    \renewcommand{\tabularxcolumn}[1]{m{#1}}

    \begin{tabularx}{\linewidth}{
        @{}
        >{\raggedright\arraybackslash}X
        >{\centering\arraybackslash}m{0.16\linewidth}
        @{}
    }
        \toprule
        \textbf{Condition} & \textbf{Weight} \\
        \midrule

        $S_p$ writes, creates, updates, or edits a file, and $S_j$ subsequently reads, loads, opens, or searches the same file.
        & $0.55$ \\

        \addlinespace[3pt]

        $S_p$ and $S_j$ share a file path but do not satisfy the above write and read relation.
        & $0.40$ \\

        \addlinespace[3pt]

        No shared file path is identified, but matching file or path tokens appear in the action summaries, action inputs, or observation summaries of $S_p$ and $S_j$.
        & $0.35$ \\

        \bottomrule
    \end{tabularx}

    \caption{Weighting rules for supplementary error propagation edges.}
    \label{tab:stage3-supplementary-edge-weights}
\end{table}


\subsection{Computation Details of Error Origin Strength $O_{ij}$}
\label{app:Oij compute}
For candidate step $S_j$, the root cause responsibility score $\operatorname{Resp}_{ij}$ is defined as
\begin{equation}
\operatorname{Resp}_{ij}
=
\min\left\{
1,\,
c_{ij}\max(O_{ij},0.05)P_i(S_j)
\right\}.
\label{eq:app:stage3-responsibility}
\end{equation}

Here, $O_{ij}$ denotes the error origin strength of candidate step $S_j$, measuring the extent to which the step itself introduces the error indicated by the user's negative feedback. It is computed as
\begin{equation}
O_{ij}
=
\widetilde a_{ij}
\left(1-h_{ij}^{\mathrm{eff}}\right).
\label{eq:stage3-origin-strength}
\end{equation}

Here, $\widetilde a_{ij}$ denotes the error introduction score and is usually set to $\widetilde a_{ij}=a_{ij}$. When the step has the role of Supporting Context or Irrelevant and its feedback relevance satisfies $r_{ij}\geq 0.4$, SIRCA instead sets $\widetilde a_{ij}=\max(a_{ij},0.2)$. The term $h_{ij}^{\mathrm{eff}}$ denotes the effective error inheritance score of candidate step $S_j$, determined by the step's own error inheritance score and the error propagation strengths from its effective predecessor steps:
\begin{equation}
h_{ij}^{\mathrm{eff}}
=
\max\left(
\{h_{ij}\}
\cup
\left\{
\eta_{ipj}
\mid
p\in\operatorname{Pred}_{ij}^{\mathrm{eff}}
\right\}
\right).
\label{eq:app:stage3-effective-inheritance}
\end{equation}
Here, $\operatorname{Pred}_{ij}^{\mathrm{eff}}$ denotes the set of effective predecessor steps of $S_j$. For a missing action hypothesis $z$, SIRCA sets $O_{iz}=a_{iz}$.


\subsection{Responsibility Score Ranking Rules for Root Cause Candidate Steps}
\label{app:candidate-eligibility-ranking}

For each negative feedback item $F_i$, SIRCA first ranks the candidates eligible for static intervention in descending order of their root cause responsibility scores $\operatorname{Resp}_i(v)$. If the scores are equal, SIRCA further compares the candidates' causal role priorities. Error Origin has the highest priority, followed by Error Propagation and Recovered Error, then Supporting Context, and finally the remaining roles. When role priorities are equal, candidates are ranked by feedback relevance $r_i(v)$ in descending order.

If the above criteria are still tied, SIRCA further ranks the candidates by their raw responsibility scores in descending order. Unlike the root cause responsibility score, the raw responsibility score does not impose a lower bound on error origin strength. For candidate step $S_j$, the raw responsibility score is computed as
\begin{equation}
\operatorname{Resp}_{ij}^{\mathrm{raw}}
=
c_{ij}O_{ij}P_i(S_j).
\end{equation}
For a missing action hypothesis $z$, the raw responsibility score is computed as
\begin{equation}
\operatorname{Resp}_{iz}^{\mathrm{raw}}
=
c_{iz}O_{iz}P_i(z).
\end{equation}

If the candidates remain tied, they are ranked by ascending step index. Since a missing action hypothesis has no step index, it is ranked ahead of executed candidate steps when all preceding criteria are tied.

\section{Computation Details of Static Intervention Combination Scores}
\label{app:intervention-combination-scoring}

This section presents the detailed computation of the metrics used to evaluate the repair effects of static intervention combinations. As in the main text, let $\mathcal U\subseteq\mathcal I_i$ denote a combination of one to three intervention proposals. The set of intervention objects involved is defined as
\begin{equation}
Z(\mathcal U)
=
\bigcup_{I_{ir}\in\mathcal U} Z_{ir}.
\label{eq:appc-joint-intervention-objects}
\end{equation}
Here, $Z_{ir}$ denotes the set of candidate objects involved in intervention proposal $I_{ir}$. Next, SIRCA removes the nodes corresponding to $Z(\mathcal U)$ and their incident edges from the attribution graph $G$ to approximate a static intervention. It then recomputes the graph path strengths from the remaining eligible candidate nodes to the negative feedback node $v_i^F$. The total graph path strength before intervention is computed as
\begin{equation}
T_i(G)
=
\sum_{v\in\mathcal H_i}
P_{i,G}^{\mathrm{graph}}(v).
\label{eq:appc-path-mass}
\end{equation}

After removing $Z(\mathcal U)$, the remaining attribution graph is denoted by $G\setminus Z(\mathcal U)$. Its total path strength is computed as
\begin{equation}
T_i\!\left(G\setminus Z(\mathcal U)\right)
=\sum_{v\in\mathcal H_i\setminus Z(\mathcal U)}
P_{i,G\setminus Z(\mathcal U)}^{\mathrm{graph}}(v).
\label{eq:appc-residual-path-mass}
\end{equation}
Here, $P_{i,G}^{\mathrm{graph}}(v)$ and
$P_{i,G\setminus Z(\mathcal U)}^{\mathrm{graph}}(v)$
denote the graph path strengths of candidate node $v$ computed before and after the intervention, respectively. Accordingly, the path blocking rate of intervention combination $\mathcal U$ is computed as
\begin{equation}
B_i\!\left(Z(\mathcal U)\right)
=
\begin{cases}
\displaystyle
1-\frac{T_i\!\left(G\setminus Z(\mathcal U)\right)}{T_i(G)},
& T_i(G)>0,\\[7pt]
0,
& T_i(G)=0.
\end{cases}
\label{eq:appc-path-break-rate}
\end{equation}
The path blocking rate primarily measures the relative decrease in the total graph path strength after the joint intervention; a larger value indicates that the intervention combination more strongly blocks error propagation paths.

Furthermore, the repair sufficiency and repair minimality of intervention combination $\mathcal U$ are computed, respectively, as
\begin{equation}
\operatorname{Suf}_i(\mathcal U)
=
1-
\prod_{I_{ir}\in\mathcal U}
\left(1-\operatorname{Suf}_{ir}\right).
\label{eq:appc-set-sufficiency}
\end{equation}

\begin{equation}
\operatorname{Min}_i(\mathcal U)
=
\frac{1}{|\mathcal U|}
\sum_{I_{ir}\in\mathcal U}
\operatorname{Min}_{ir}.
\label{eq:appc-set-minimality}
\end{equation}

After obtaining the path blocking rate $B_i\!\left(Z(\mathcal U)\right)$, repair sufficiency $\operatorname{Suf}_i(\mathcal U)$, and repair minimality $\operatorname{Min}_i(\mathcal U)$, SIRCA can compute the static counterfactual repair score of intervention combination $\mathcal U$ to comprehensively evaluate its expected repair effect on the issue indicated by the current negative feedback, as follows:
\begin{equation}
\begin{aligned}
\operatorname{Score}_{\mathrm{CF},i}(\mathcal U)
={}&0.50B_i\!\left(Z(\mathcal U)\right)
+0.35\operatorname{Suf}_i(\mathcal U)\\
&+0.15\operatorname{Min}_i(\mathcal U)
-\operatorname{Pen}(\mathcal U).
\end{aligned}
\end{equation}

However, the static counterfactual repair score $\operatorname{Score}_{\mathrm{CF},i}(\mathcal U)$ cannot directly reflect the attribution responsibility of the candidate objects in the intervention combination. Therefore, SIRCA combines it with the relative responsibility strength $\operatorname{Resp}_i^{\mathrm{rel}}(\mathcal U)$ of the combination to compute the final attribution score:
\begin{equation}
\begin{aligned}
\operatorname{Score}_{\mathrm{FA},i}(\mathcal U)
={}&0.65\operatorname{Score}_{\mathrm{CF},i}(\mathcal U)\\
&+0.35\operatorname{Resp}_i^{\mathrm{rel}}(\mathcal U).
\end{aligned}
\end{equation}

Here, $\operatorname{Resp}_i^{\mathrm{rel}}(\mathcal U)$ measures the ratio of the highest responsibility score among the candidate objects in the intervention combination to the highest responsibility score among all eligible candidate objects for the current negative feedback. It is computed as
\begin{equation}
\operatorname{Resp}_i^{\mathrm{rel}}(\mathcal U)
=
\frac{
\displaystyle\max_{v\in Z(\mathcal U)}\operatorname{Resp}_i(v)
}{
\displaystyle\max\!\left(
\max_{v\in\mathcal H_i}\operatorname{Resp}_i(v),
\varepsilon
\right)
}.
\label{eq:appc-relative-responsibility}
\end{equation}
Here, $Z(\mathcal U)\subseteq\mathcal H_i$, and $\operatorname{Resp}_i(v)$ denotes the root cause responsibility score of candidate object $v$; $\varepsilon=10^{-9}$ is used to avoid division by zero.

\section{Single Root Cause Output Adaptation}
\label{app:public-trajectory-adaptation}

The root cause set produced by SIRCA may contain multiple steps and thus does not directly satisfy the evaluation requirement of identifying a single decisive error step in single root cause localization tasks. To support such tasks, we optionally add a fifth stage after SIRCA's attribution results, which ultimately outputs a single decisive error step $S_{\hat{j}}$. This stage is introduced only to adapt SIRCA to the evaluation requirements of single root cause localization tasks and is not part of SIRCA's core attribution pipeline.

We assume that single root cause localization tasks, such as Who\&When, provide the task description $q$, the reference answer $y^*$, and task failure information required for subsequent single root cause output adaptation.

\subsection{Initial Decisive Error Localization}

Let $q$ denote the task description, $y^{*}$ the reference answer,
and $\mathcal S$ the step sequence corresponding to the trajectory's
original interaction history:
\begin{equation}
\mathcal S=(S_0,\ldots,S_{N-1}).
\end{equation}

Inspired by the Step by Step strategy
\cite{zhang2025agent}, SIRCA analyzes the trajectory
$\mathcal{S}$ in chronological order to assess failures at each step. For the current step $S_j$, the LLM-based judge $\mathcal{J}_{\mathrm{SGS}}$ considers the task description $q$, the reference answer $y^{*}$, and the interaction history $\mathcal{S}_{0:j}$ up to $S_j$ to determine whether $S_j$ is the first decisive error step leading to task failure.

SIRCA also provides $\mathcal{J}_{\mathrm{SGS}}$ with the structural information $c_j$ of the current step as a soft prior for this judgment:
\begin{equation}
c_j=
\left(
\ell_{ij},
a_{ij},
\chi_{ij}^{\mathrm{root}},
\mathbb{I}[S_j\in\mathcal{C}_i^{*}]
\right).
\end{equation}
Here, $i$ indexes the feedback item $F_i$ considered in the current attribution; $\ell_{ij}$ denotes the step role; $a_{ij}$ denotes the error introduction score; 
$\chi_{ij}^{\mathrm{root}}$ indicates whether step $S_j$ satisfies the static intervention eligibility criteria, taking a value of $1$ if the criteria are met and $0$ otherwise; $\mathcal{C}_i^{*}$ denotes the final root cause set. The structural information serves only as an auxiliary reference and does not directly determine the localization result; $\mathcal{J}_{\mathrm{SGS}}$ still makes its judgment based on the task requirements, the reference answer, and the actual trajectory content.

SIRCA stops scanning upon obtaining the first valid affirmative judgment and takes the corresponding step as the initial localization result, denoting its step index by $\hat{j}_{\mathrm{SGS}}$.

\subsection{Root Cause Step Correction}

After obtaining the initial localization result $S_{\hat{j}_{\mathrm{SGS}}}$, SIRCA further invokes an LLM-based decision model $\mathcal{J}$ to revise it, thereby avoiding misidentifying steps that merely propagate or manifest errors as the steps that first introduced them.

Specifically, SIRCA takes the initial localization result $S_{\hat{j}_{\mathrm{SGS}}}$, the complete task failure information, and the trajectory content and attribution relations of each step in the candidate set $\mathcal{K}$ as inputs to $\mathcal{J}$. The decision model comprehensively compares the candidate steps and selects the step that better explains the error origin and the task failure as the final localization result.

The candidate step set is defined as:
\begin{equation}
\mathcal{K}
=
\operatorname{Select}\!\left(
\mathcal{K}^{(3)}
\cup
\mathcal{K}^{(4)}
\cup
\mathcal{K}^{\mathrm{fail}}
\cup
\mathcal{K}^{\mathrm{adj}}
\right).
\end{equation}

Here, $\mathcal{K}^{(3)}$ denotes the candidate step set obtained during attribution graph construction, $\mathcal{K}^{(4)}$ denotes the candidate step set obtained during static intervention, and $\mathcal{K}^{\mathrm{fail}}$ denotes the set of relevant steps retrieved based on task failure manifestations. The set $\mathcal{K}^{\mathrm{adj}}$ denotes the set containing the initial localization step and its immediately preceding and following steps, defined as:
\begin{equation}
\mathcal{K}^{\mathrm{adj}}
=
\left\{
S_k
\;\middle|\;
k\in\{0,\ldots,N-1\},
\ |k-\hat{j}_{\mathrm{SGS}}|\leq 1
\right\}.
\end{equation}

The function $\operatorname{Select}(\cdot)$ is used to merge, deduplicate, and filter candidate steps from different stages. Finally, the decision model $\mathcal{J}$ outputs the corrected decisive error step index $\hat{j}_{\mathrm{CFS}}$.

In addition to the above corrections, some trajectory datasets, such as Who\&When, may contain consecutive operations such as searches, page navigation, and state updates, causing $S_{\hat j_{\mathrm{CFS}}}$ to correspond to a later operation rather than a related earlier step. SIRCA therefore performs deterministic structural correction based on action types, temporal order, and content associations in the trajectory. For example, if $S_{\hat j_{\mathrm{CFS}}}$ corresponds to a scrolling step late in the trajectory while a search step occurs earlier, SIRCA may move the localization result back to that search step based on their action types and temporal order. SIRCA adjusts the result only when the corresponding structural rules provide sufficient grounds; otherwise, it retains $S_{\hat j_{\mathrm{CFS}}}$.

The final decisive error step is defined as
\begin{equation}
S_{\hat j}
=
\Phi_{\mathrm{SC}}
\left(
S_{\hat j_{\mathrm{CFS}}},
\mathcal S
\right),
\label{eq:stage5-structure-correction}
\end{equation}
where $\Phi_{\mathrm{SC}}$ denotes the correction function based on the trajectory structure. After this correction, SIRCA ultimately produces the decisive error step $S_{\hat j}$ required for evaluating single root cause localization tasks.

\section{Root Cause Localization Metrics}

\label{app:evaluation-metrics}

This section introduces the evaluation metrics for root cause localization accuracy across the datasets. Let the 
evaluation dataset be

\begin{equation}
    \mathcal{D}
    =
    \{\tau_i\}_{i=1}^{N},
    \label{eq:evaluation-dataset}
\end{equation}
where $N$ denotes the number of trajectory samples in the evaluation dataset. 

For each trajectory sample $\tau_i \in \mathcal{D}$, let $\mathcal{S}_i^{\star}$ denote the set of ground truth root cause steps. 
The deduplicated set of root cause steps predicted by a given method for trajectory $\tau_i$ is denoted by $\widehat{\mathcal{S}}_i$. For a trajectory sample $\tau_i$, let $m_i$ denote the number of root cause steps correctly identified in this trajectory, computed as
\begin{equation}
m_i
=
\left|
\mathcal{S}_i^{\star}
\cap
\widehat{\mathcal{S}}_i
\right|.
\label{eq:correctly-identified-root-causes}
\end{equation}

\subsection{Evaluation Metrics for the Multi Root Cause Dataset}

\subsubsection{Trajectory Averaged Root Cause Precision}

Trajectory averaged root cause precision $\mathrm{Traj\text{-}P}$ measures the proportion of predicted root cause steps that are ground truth root cause steps. For a trajectory sample $\tau_i$, its precision is defined as:
\begin{equation}
P_i=
\begin{cases}
\dfrac{m_i}{|\widehat{\mathcal S}_i|}, & |\widehat{\mathcal S}_i|>0,\\[6pt]
0, & |\widehat{\mathcal S}_i|=0.
\end{cases}
\end{equation}

Here, $|\widehat{\mathcal S}_i|$ denotes the number of deduplicated root cause steps predicted by the given method for trajectory sample $\tau_i$.

Finally, the arithmetic mean of precision across all trajectory samples gives the trajectory averaged root cause precision:
\begin{equation}
\mathrm{Traj\text{-}P}
=
\frac{1}{N}\sum_{i=1}^{N}P_i.
\end{equation}

\subsubsection{Trajectory Averaged Root Cause Recall}

The trajectory averaged root cause recall $\mathrm{Traj\text{-}R}$ measures the proportion of ground truth root cause steps predicted by different methods among all annotated root cause steps. For trajectory $\tau_i$, its recall is defined as
\begin{equation}
    R_i=
\begin{cases}
\dfrac{m_i}{|\mathcal S_i^{\star}|}, & |\mathcal S_i^{\star}|>0,\\[6pt]
0, & |\mathcal S_i^{\star}|=0.
\end{cases}
\end{equation}

Here, $|\mathcal S_i^{\star}|$ denotes the number of ground truth root cause steps annotated for trajectory $\tau_i$. Averaging over all trajectories gives
\begin{equation}
    \mathrm{Traj\text{-}R}
=
\frac{1}{N}\sum_{i=1}^{N}R_i.
\end{equation}

\subsubsection{Trajectory Averaged Root Cause $F_1$ Score}
The trajectory averaged root cause $F_1$ Score $\mathrm{Traj\text{-}F}_1$ evaluates the combined performance of different methods in prediction accuracy and coverage of ground truth root causes. For trajectory $\tau_i$, its $F_1$ score is defined as
\begin{equation}
    F_{1,i}=
\begin{cases}
\dfrac{2P_iR_i}{P_i+R_i}, & P_i+R_i>0,\\[6pt]
0, & P_i+R_i=0.
\end{cases}
\end{equation}

Averaging over all trajectories gives
\begin{equation}
  \mathrm{Traj\text{-}F}_1
=
\frac{1}{N}\sum_{i=1}^{N}F_{1,i}.  
\end{equation}

\subsubsection{Any Root Cause Step Hit Rate}

The any root cause step hit rate $\mathrm{Hit}_{\mathrm{step}}^{\mathrm{any}}$ measures whether the predictions of a given method identify at least one annotated ground truth root cause step. For a trajectory sample $\tau_i$, its hit status is defined as:
\begin{equation}
H_i^{\mathrm{any}}=
\begin{cases}
1, & m_i>0,\\
0, & m_i=0.
\end{cases}
\end{equation}

Here, $H_i^{\mathrm{any}}$ indicates whether trajectory sample $\tau_i$ contains at least one correctly identified root cause step.

Averaging the hit indicators over all trajectory samples gives the any root cause step hit rate:
\begin{equation}
\mathrm{Hit}_{\mathrm{step}}^{\mathrm{any}}
=
\frac{1}{N}\sum_{i=1}^{N}H_i^{\mathrm{any}}.
\end{equation}

\subsection{Evaluation Metrics for the Single Root Cause Dataset}

\subsubsection{Exact Root Cause Localization Accuracy}

Exact root cause localization accuracy $\mathrm{Acc}_{\mathrm{step}}^{\mathrm{exact}}$ measures whether the decisive error step predicted by a given method exactly matches the annotated step, thereby reflecting its root cause localization accuracy.

For a trajectory sample $\tau_i$ in the dataset, let $\hat{j}_i$ and $j_i^{\star}$ denote the predicted and annotated decisive error step indices, respectively. The exact localization outcome for this trajectory sample is defined as:
\begin{equation}
z_i^{\mathrm{exact}}
=
\begin{cases}
1, & \hat{j}_i\neq\varnothing \land \hat{j}_i=j_i^{\star},\\[6pt]
0, & \text{otherwise}.
\end{cases}
\label{eq:exact-localization-outcome}
\end{equation}

Averaging over all trajectory samples in the dataset gives the exact root cause localization accuracy:
\begin{equation}
\mathrm{Acc}_{\mathrm{step}}^{\mathrm{exact}}
=
\frac{1}{N}\sum_{i=1}^{N}z_i^{\mathrm{exact}}.
\label{eq:exact-step-accuracy}
\end{equation}

\subsubsection{Tolerance Aware Root Cause Localization Accuracy}

Tolerance aware root cause localization accuracy, denoted by
$\mathrm{Acc}_{\mathrm{step}}^{(\pm1)}$, measures how accurately a given method localizes the decisive error step when a localization deviation of at most one step is allowed.

For a trajectory sample $\tau_i$ in the dataset, its localization outcome under this tolerance is defined as:
\begin{equation}
z_i^{(\pm1)}
=
\begin{cases}
1, & \hat j_i\neq\varnothing \land
     |\hat j_i-j_i^{\star}|\leq 1,\\[6pt]
0, & \text{otherwise}.
\end{cases}
\label{eq:tolerant-localization-outcome}
\end{equation}

Here, $|\hat j_i-j_i^{\star}|$ denotes the difference between the predicted step index and the annotated step index.

Averaging over all trajectory samples in the dataset gives the tolerance aware root cause localization accuracy:
\begin{equation}
\mathrm{Acc}_{\mathrm{step}}^{(\pm1)}
=
\frac{1}{N}\sum_{i=1}^{N}z_i^{(\pm1)}.
\label{eq:tolerant-step-accuracy}
\end{equation}

\section*{Declaration of Generative AI Use}

We used Generative AI to assist with language polishing and code writing. All AI-generated suggestions and code were carefully reviewed, tested, and verified by the authors, who take full responsibility for the final manuscript and code.

\bibliographystyle{named}
\bibliography{ijcai27}

\end{document}